\documentclass[10pt,twocolumn,letterpaper]{article}
\usepackage{cvpr}              

\usepackage{graphicx}
\usepackage{amsmath}
\usepackage{amssymb}
\usepackage{booktabs}
\usepackage{multirow}
\usepackage{enumitem}
\usepackage{indentfirst}  

\usepackage[T1]{fontenc}
\usepackage[utf8]{inputenc}
\usepackage[final]{microtype}

\DeclareUnicodeCharacter{2011}{-}
\DeclareUnicodeCharacter{00AD}{}
\DeclareUnicodeCharacter{00A0}{\nobreakspace{}}
\DeclareUnicodeCharacter{2013}{--}
\DeclareUnicodeCharacter{2014}{---}
\DeclareUnicodeCharacter{2018}{`}
\DeclareUnicodeCharacter{2019}{'}
\DeclareUnicodeCharacter{201C}{``}
\DeclareUnicodeCharacter{201D}{''}
\DeclareUnicodeCharacter{2248}{$\approx$}
\DeclareUnicodeCharacter{00D7}{$\times$}
\DeclareUnicodeCharacter{2194}{$\leftrightarrow$}

\newcommand{\faceocc}{\text{face-occ}}
\newcommand{\bgocc}{\text{bg-occ}}
\newcommand{\supp}{the supplementary material}

\definecolor{cvprblue}{rgb}{0.21,0.49,0.74}
\usepackage[pagebackref,breaklinks,colorlinks,allcolors=cvprblue]{hyperref}
\usepackage[accsupp]{axessibility}  

\def\paperID{24226} 
\def\confName{CVPR}
\def\confYear{2026}

\title{From Measurement to Mitigation: Quantifying and Reducing Identity Leakage in Image Representation Encoders with Linear Subspace Removal}

\author{Daniel George, Charles Yeh, Daniel Lee, Yifei Zhang \\
Persona Identities, USA\\
\texttt{\{daniel.george,charles,daniel,yifei\}@withpersona.com}
}

\begin{document}
\maketitle
\begin{abstract}
    Frozen visual embeddings (e.g., CLIP, DINOv2/v3, SSCD) power retrieval and integrity systems, yet their use on face-containing data is constrained by unmeasured identity leakage and a lack of deployable mitigations. We take an attacker-aware view and contribute: (i) a benchmark of visual embeddings that reports open-set verification at low false-accept rates, a calibrated diffusion-based template inversion check, and face–context attribution with equal-area perturbations; and (ii) propose a one-shot linear projector that removes an estimated identity subspace while preserving the complementary space needed for utility, which for brevity we denote as the identity sanitization projection ISP. Across CelebA-20 and VGGFace2, we show that these encoders are robust under open-set linear probes, with CLIP exhibiting relatively higher leakage than DINOv2/v3 and SSCD, robust to template inversion, and are context-dominant. In addition, we show that ISP drives linear access to near-chance while retaining high non-biometric utility, and transfers across datasets with minor degradation. Our results establish the first attacker-calibrated facial privacy audit of non-FR encoders and demonstrate that linear subspace removal achieves strong privacy guarantees while preserving utility for visual search and retrieval.
\end{abstract}

\section{Introduction}

\begin{figure}[t]
    \centering
    \includegraphics[width=\linewidth]{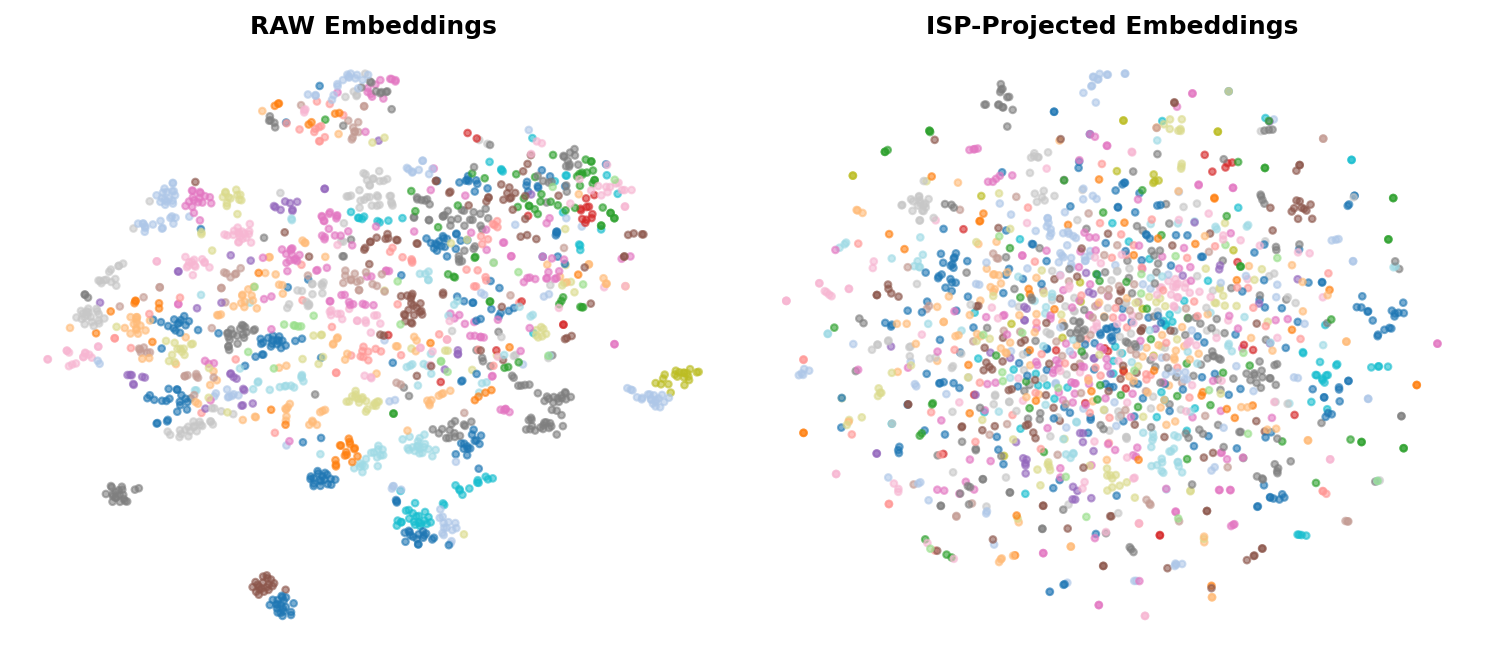}
    \caption{%
        \textbf{Identity Clustering in CLIP Embeddings.}
        t-SNE visualization (perplexity=30, 1000 iterations) of test embeddings
        colored by identity. Left: RAW embeddings exhibit visible identity clustering.
        Right: ISP-sanitized embeddings with identity structure removed via
        orthogonal projection onto complement of top-192 identity directions.
        Embeddings were calculated over a randomly sampled VGGFace2 test split of 80 identities and 1,600 images (20 per identity).
        t-SNE computed on L2-normalized 512-dimension embeddings.
    }
    \label{fig:tsne_clip}
\end{figure}

Embedding-based features and vector search underpin large-scale retrieval and integrity systems (near-duplicate search, portrait reuse, copy/manipulation detection). In practice, teams often deploy frozen (or lightly tuned) encoders with ANN indices for stability and throughput. While face recognition~(FR) systems are intentionally biometric and operate at very low false-accept rates (FAR), other widely used encoders (e.g., CLIP, DINOv2/v3, SSCD) are trained without identity supervision and are repurposed for non-biometric tasks.  When such encoders are applied to face-containing data, operators face a practical trade-off: the invariances that make these features robust for search and integrity can also expose residual biometric cues. Because identity leakage in non-FR encoders is rarely measured at open-set, low-FAR operating points (and for DINOv2, DINOv3, and SSCD, measured at all), deployments lack a calibrated way to certify “safe” use.

Our work targets this measurement gap through an attacker-aware lens. First, we evaluate DINOv2, DINOv3, SSCD, and CLIP using a measurement suite for frozen encoders that (i) quantifies open-set verification at low FAR with strong linear probes, (ii) audits calibrated template inversion with a diffusion prior, and (iii) localizes \emph{where} identity lives via face–context attribution with equal-area perturbations. Second, we propose \textbf{Identity Sanitization Projection (ISP)}, a lightweight post-hoc projector that removes an estimated identity subspace, and show that it lowers susceptibility to adversarial attacks while preserving embedding utility heavily.

Across CLIP~\cite{Radford2021}, DINOv2~\cite{Oquab2023}, DINOv3~\cite{Oquab2025DINOv3}, and SSCD~\cite{Pizzi2022SSCD}, we find that these encoders exhibit minimal linear identity accessibility at low FAR; ISP suppresses this linear access to near-chance while maintaining high non‑biometric utility. To our knowledge, prior audits either target FR models or study CLIP in isolation; we provide the first attacker-calibrated, open-set, low-FAR evaluation for DINOv2/v3 and SSCD alongside CLIP, and release projectors/diagnostics for adoption.

\textbf{Contributions.} (1) An attacker-calibrated audit for facial privacy in visual encoders that reports open-set true acceptance rate (TAR) at low-FAR, calibrated template inversion, and face–context attribution. (2) A one‑shot, moment‑based projector for facial identity. (3) Evidence that the estimated identity subspace is compact and transferable across datasets, enabling a fixed, auditable $P$ that suppresses open‑set linear access while preserving utility. In addition, we intend to open source our codebase, including our projections and our evaluation toolkit.

\section{Background and Related Work}
\label{sec:related}

\paragraph{Deployment context and privacy constraints.}
Many online services—including financial institutions, gig economy platforms, e-commerce marketplaces, travel and rental companies, and social networks—require robust identity verification and fraud prevention to protect both users and their businesses~\cite{Zhang2025IdentityFraud}. In these scenarios, maintaining trust and safety depends on the ability to detect duplicate or manipulated identity documents, as well as AI-generated or face-swapped images~\cite{Papantoniou2024}. Organizations increasingly require tools that can identify images with similar backgrounds or other shared visual elements to combat such fraud. However, many are unable to use biometric technologies such as facial recognition or fingerprint matching because of strict privacy regulations (e.g., GDPR, CCPA), industry-specific compliance requirements, or internal policies aimed at protecting user privacy and minimizing legal risk~\cite{Fredrikson2015}. In some sectors, the use of biometrics is explicitly restricted or requires additional user consent, data protection measures, and regulatory oversight, making implementation complex and costly. This regulatory landscape creates demand for non-FR encoders that can perform integrity and similarity tasks \emph{without} being explicitly biometric—yet the privacy properties of such encoders remain poorly characterized under adversarial threat models~\cite{Hintersdorf2024,Caldarella2024}.

\paragraph{Visual encoders.}
Modern practice for frozen visual features centers on three families: (i) self-supervised ViT encoders (e.g., DINO/iBOT/MAE), with DINOv2 as a widely used scaled exemplar \cite{Oquab2023}; (ii) vision--language encoders (CLIP/ALIGN) that align images and text at web scale \cite{Radford2021,Jia2021}; and (iii) copy-detection descriptors designed for manipulation/near-duplicate robustness (SSCD) \cite{Pizzi2022SSCD}. Despite their ubiquity, there is limited \emph{attacker-aware} evidence about whether these embeddings are privacy-preserving for faces under \emph{open-set}, \emph{low–false-accept} operation. Interpretability efforts help characterize what these models represent--via saliency/perturbation/attention analyses \cite{Zeiler2014,Fong2017,Petsiuk2018,Chefer2021} and, more recently, geometric/mechanistic views that argue concepts may behave like archetypes or regions rather than single directions \cite{Fel2025RabbitHull}.

However, these tools are not calibrated to biometric operating points, and that gap is precisely where facial privacy risk lives at scale. In real deployments, operators target FAR in the $10^{-4}$–$10^{-6}$ range (FRVT/ISO guidance) to keep the total number of impostor accepts manageable across billions of comparisons \cite{Grother2019,ISO19795}. Saliency maps and concept directions do not yield an operating threshold or a measured error rate; they reason about \emph{single images} (or classifier logits) rather than pairwise similarity decisions in a frozen encoder, and they rarely enforce identity-disjoint evaluation. As a result, they cannot answer the attacker‑relevant question our paper centers on: \emph{is identity information linearly accessible at low FAR for unseen identities, and if so, how can we reduce this effect while still preserving embedding utility?}
To our knowledge, no prior work has provided such an attacker-calibrated audit for DINOv2, DINOv3, or SSCD-- all analyses focus on CLIP or on dedicated FR embeddings and report closed-set accuracy or aggregate leakage signals without low-FAR calibration \cite{Hintersdorf2024,Caldarella2024}. This motivates our evaluation suite: open-set, low-FAR verification probes that deliver TAR@FAR metrics, a calibrated template-inversion protocol, and attribution diagnostics tailored to pairwise similarity. Building on these measurements, we introduce a low-latency, auditable, null-space projection (ISP)--that \emph{certifies} linear removal of identity-aligned directions while preserving non-biometric utility.

\paragraph{Subspace removal.}
Removing sensitive attributes from fixed representations via linear editing falls into two families: \emph{classifier-driven} methods (INLP, RLACE~\cite{Ravfogel2020,Ravfogel2022RLACE}) that iteratively train linear adversaries and project onto their nullspaces, and \emph{moment-based} methods (SAL, LEACE~\cite{Shao2023SAL,Belrose2023LEACE}) that remove between-class mean structure in a single closed-form step. The former offer strong linear unpredictability guarantees but require repeated fitting and are difficult to audit; the latter are one-shot and low-distortion but have been applied primarily to binary or low-cardinality attributes.

Facial identity in our setting is high-cardinality and open-set, and risk must be quantified at low FAR under disjoint identities. It is therefore unclear a priori whether the identity signal in non-FR embeddings (a)~concentrates in a compact, transferable subspace, (b)~can be linearly removed without harming non-biometric utility, and (c)~remains suppressed under non-linear probes. We answer these questions with attacker-aware metrics (open-set TAR@low-FAR, projection-only MLP, template inversion) and cross-dataset transfer checks that are typically absent from concept-erasure evaluations. Our moment-based projector, ISP, is detailed in \S\ref{sec:ISP}; extended background on subspace removal methods can be found in Appendix~\ref{app:subspace-removal}.
\paragraph{Face--Context Attribution in Visual Embeddings.}
Context bias and shortcut reliance are well studied in object recognition~\cite{Geirhos2019,Xiao2020} and vision--language models~\cite{Li2023CLIPExplain,Wu2025SegDebias}, and explainability methods for FR focus on single-image saliency via occlusion or perturbation maps~\cite{Fong2017,Petsiuk2018,Mery2022MinPlus,Adhikari2024LostInContext}. However, these tools reason about individual images or class logits rather than pairwise similarity in frozen embeddings, and do not quantify face vs.\ context importance at biometric operating points.

We introduce three complementary diagnostics for pairwise face--context attribution: \emph{FII}, which compares equal-area face vs.\ background occlusions; \emph{$B^*$}, a stress test measuring how much revealed background is needed for context to overtake identity; and \emph{CPI}, a face-blur sweep tracking identity-vs-context preference as facial detail degrades. All perturbations are area-normalized via a common face coverage ratio (FCR) to ensure commensurate comparisons. Formal definitions and protocol details are in \supp.

\paragraph{Template inversion.} A separate line of work reconstructs faces directly from stored templates, demonstrating that modern FR embeddings carry a strong, prior–navigable identity signal \cite{Fredrikson2015,Mahendran2015,Mai2017,Shahreza2023,Struppek2022,Wang2025,Papantoniou2024}. However, nearly all evaluations target \emph{FR models} and report success without grounding to open–set, low–FAR operation in which deployments actually run. This leaves a hole for visual encoders: linear verification probes quantify decision–boundary accessibility, but they do not test \emph{generative recoverability}--whether a strong diffusion prior can traverse the embedding’s identity footprint to synthesize a face that cross–verifies at deployment thresholds. We therefore include a budget–matched diffusion inversion audit and judge success via \emph{cross–model} FR verification (Sec.~\ref{sec:r5}). This provides an attacker–aware complement to our linear experiments. Low TAR@low–FAR alongside failed inversion offers stronger evidence of limited identity access than either signal alone, while any inversion success would reveal non–linear/generative leakage that linear erasure cannot preclude. It must also be noted, however, that inversion outcomes are prior– and budget–dependent, so negative results are not proofs of privacy. \bigskip

\section{Methodology}
\label{sec:methodology}

In this paper, we run our experiments over datasets with labeled identities. For each image dataset $I$, we construct a subset $C \subseteq I$ such that each individual in the dataset has exactly $n$ images, and there are $m$ individuals in the subset. Let $A := \{1,\ldots,m\}$ be the set of identities and $B_i$ be the set of images for identity $i \in A$, such that $|B_i|=n$ \text{ for all } $i \in A$. Let $f:\mathcal{X}\!\to\!\mathbb{R}^d$ be a frozen encoder and
\begin{equation}
    z \;=\; \frac{f(x)}{||f(x)||_2}\in\mathbb{R}^d
\end{equation}
the $\ell_2$-normalized embedding of image $x$. Thus for any identity $i \in A$, we have $n$ normalized embeddings
\begin{equation}
    z_i^j \quad \text{for all } j \in \{1, \ldots, n\}.
\end{equation}

\subsection{Few-shot probing}
\label{sec:probing}

We evaluate identity accessibility via open-set, few-shot verification.

For each identity set $B_i$, we split its $n$ embeddings into a \emph{support set} $S_i$ (used for training) and a \emph{query set} $Q_i$ (held out for evaluation), with $S_i \cap Q_i = \varnothing$.
The verifier learns a projection $W\!\in\!\mathbb{R}^{d\times r}$ that maps each embedding individually into a lower-dimensional space.
Given a query–support pair $(z_q,\, z_s)$, the verification score is the cosine similarity of their projections:
\begin{equation}
    \text{score}(z_q, z_s) \;=\;
    \frac{(z_q W)^\top (z_s W)}{\|z_q W\|_2\,\|z_s W\|_2},
\end{equation}
and the pair is accepted as a match if $\text{score}(z_q, z_s) \geq \tau$.
$W$ is trained on mated ($y{=}1$, same identity) and impostor ($y{=}0$, different identity) pairs from $A_{\text{train}}$.
We evaluate two probe families: \emph{Ridge}, which learns $W$ via $\ell_2$-regularised least squares, and \emph{MLP}, which learns a non-linear projection via a two-layer network trained with cross-entropy. $k$ denotes the number of support embeddings per identity available during training.

We adopt an \emph{open-set, identity-disjoint} protocol: identities are partitioned into train/val/test splits ($A_{\text{train}} / A_{\text{val}} / A_{\text{test}}$) with no overlap. A ridge regression verifier is trained on $A_{\text{train}}$; all hyperparameters and a verification threshold $\tau$ are selected on $A_{\text{val}}$ and frozen before evaluation on $A_{\text{test}}$. A separate $\tau$ is calibrated for each (dataset, encoder, projection, $k$) combination; once set on validation identities, it is held fixed for all test resamplings. We target $\mathrm{FAR} \approx 10^{-4}$ where impostor counts permit, and report $\mathrm{TAR}$ on test identities at $k \in \{1, 4, 16\}$ to probe few-shot through many-shot attacker regimes. Results are averaged over five seeds that resample $S_i / Q_i$ per identity, with 95\% identity-aware confidence intervals. Full details on probe inputs, pair construction, threshold calibration, FAR denominators, and per-seed logs are in Appendix~\ref{app:probing}.
\subsection{Attribution with FCR normalization}

We want to localize where \emph{identity} evidence comes from in an encoder: the face region vs.\ the surrounding context.
To make that attribution fair across heterogeneous images (tight crops vs.\ wide scenes) and robust to nuisance factors, we (i) equalize the face area across images, then (ii) apply \emph{matched} perturbations to face and background, and finally (iii) probe how preference shifts as we systematically reduce face information or reveal more background.

\paragraph{Face-coverage ratio (FCR).}
Faces occupy different fractions of an image; naively "masking the face" in a tight crop removes more pixels (and thus more evidence) than in a wide shot.
We therefore standardize the \emph{face budget} before any perturbation by targeting a common
\begin{equation}
    \mathrm{FCR}(x)=\frac{\mathrm{area}(\text{face mask in }x)}{\mathrm{area}(x)}.
\end{equation}
After resizing/cropping to reach a target FCR, we only compare perturbations that are \emph{equal-area} and \emph{equal-strength}; this lets us attribute any similarity change to \emph{what we touched} (face vs.\ background), not to \emph{how much} we touched.
(Implementation details---including mask construction and area tolerance---are in Appendix~\ref{app:fca}.)

\paragraph{Face Importance Index (FII).}

If equal-area occlusions to face and background have different effects on similarity, quantification of this difference is informative about where identity lives.
Let $z^{\bgocc}$ and $z^{\faceocc}$ be the $\ell_2$-normalized embeddings of image $x$ with an equal-area occlusion mask applied to the background region and face respectively.

For a query and reference pair, we then measure facial vs.\ background contributions using cosine similarity:

\begin{align}
    \Delta_{\text{face}} & =
    \frac{z_q^\top z_r}{\|z_q\|_2 \|z_r\|_2}
    - \frac{\big(z_q^{\faceocc}\big)^\top \big(z_r^{\faceocc}\big)}{\|z_q^{\faceocc}\|_2 \|z_r^{\faceocc}\|_2}, \\
    \Delta_{\text{bg}}   & =
    \frac{z_q^\top z_r}{\|z_q\|_2 \|z_r\|_2}
    - \frac{\big(z_q^{\bgocc}\big)^\top \big(z_r^{\bgocc}\big)}{\|z_q^{\bgocc}\|_2 \|z_r^{\bgocc}\|_2},
\end{align}
and $\mathrm{FII}=\Delta_{\text{face}}-\Delta_{\text{bg}}$ (averaged over pairs).

\paragraph{Context Preference Index (CPI).}

For each image $i$, we construct two reference images: an \emph{identity-matched} image $z_{\text{id}}^{i}$ (same person, different context) and a \emph{context-matched} image $z_{\text{ctx}}^{i}$ (different person via face-swap, same context). We apply Gaussian face blur with strength $\sigma$ to all three images and compute their embeddings. CPI measures how often the blurred query prefers context over identity:
\begin{align}
    \mathrm{CPI}(\sigma) & = \tfrac{1}{N}\!\sum_{i=1}^N
    \mathbf{1}\!\Big[\frac{z_{q,\sigma}^{i\top} z_{\text{ctx},\sigma}^{i}}{\|z_{q,\sigma}^{i}\|_2 \|z_{\text{ctx},\sigma}^{i}\|_2} \notag \\
                         & \ge
    \frac{z_{q,\sigma}^{i\top} z_{\text{id},\sigma}^{i}}{\|z_{q,\sigma}^{i}\|_2 \|z_{\text{id},\sigma}^{i}\|_2}\Big],
\end{align}
and we report $\Delta\mathrm{CPI}=\mathrm{CPI}(\sigma_{\max})-\mathrm{CPI}(\sigma_{\min})$ and the crossover $\sigma^*$ when defined.

\paragraph{Background revelation threshold ($B^*$).}
Finally, as a complementary \emph{stress test}, we monotonically reveal more background while keeping the face crop constant and ask when the \emph{context-matched} image overtakes the \emph{identity-matched} image.
Let $B\in[0,1]$ be the revealed background fraction in a monotone series that preserves the face crop.
For each triplet (query, identity-matched, context-matched), define
\begin{align}
    B^*_{\text{query}} & =\min\Big\{B:\ \frac{z_{q,B}^{\top} z_{\text{ctx}}}{\|z_{q,B}\|_2 \|z_{\text{ctx}}\|_2} \notag \\
                       & \ge \frac{z_{q,B}^{\top} z_{\text{id}}}{\|z_{q,B}\|_2 \|z_{\text{id}}\|_2}\Big\},
\end{align}
and report the median across triplets (with censoring at $0$ and $1$).
Implementation details are in Appendix~\ref{app:fca}.

\subsection{Template Inversion}
\label{sec:inversion}

Template inversion aims to produce a face image $\hat{x}$ from a target embedding $z_{\text{tgt}}$ alone, without access to the true face.
Attacks differ in their generative approach - DiffMI~\cite{Wang2025} optimizes over a face-DDPM latent pool via APGD; ALSUV~\cite{Jung2024ALSUV} searches a StyleGAN2 latent space via Adam; Vec2Face~\cite{Wu2025Vec2Face} uses a trained direct-regression generator; and Bob~\cite{Kim2024Scores} algebraically inverts an embedding score vector.

We evaluate inversion success for all attacks via cross-model face verification: embed $\hat{x}$ and the true target $x_{\text{tgt}}$ with a disjoint FR encoder $f_{\text{FR}}$ and accept if
\begin{equation}
    \frac{f_{\text{FR}}(\hat{x})^\top f_{\text{FR}}(x_{\text{tgt}})}{\|f_{\text{FR}}(\hat{x})\|_2 \|f_{\text{FR}}(x_{\text{tgt}})\|_2} \;\ge\; \tau_F,
\end{equation}
where $\tau_F$ is calibrated at minimum EER (ArcFace: $\tau_F{=}0.1051$, AdaFace: $\tau_F{=}0.1111$). Full per-attack protocol details are given in Appendix~\ref{app:inversion}.

\subsection{Identity Sanitization Projection (ISP)}
\label{sec:ISP}

Under a standard homoscedastic model where identities share a within‑class covariance $\Sigma_w$, the Bayes‑optimal linear discriminants live in the Fisher/Mahalanobis geometry induced by $\Sigma_w^{-1}$.
In that geometry, linear separability across identities is governed by the between‑class mean subspace:
if $M=[\mu_i-\mu_C]\in\mathbb{R}^{d\times m}$ stacks centered identity means, then the discriminative directions are the left singular vectors of the \emph{whitened} matrix $\tilde{M}=\Sigma_w^{-1/2}M$.
Projecting onto the orthogonal complement of the top-$r$ singular vectors removes between‑class mean structure, so any linear verifier $w^\top z$ loses its identity margin at test time:
$w^\top(\mu_i-\mu_j)\approx 0$ for all $i\neq j$.
This gives us an auditable knob ($r$) to tune privacy versus utility and a certificate against the specific attacker we evaluate—open‑set linear verification at low FAR (Sec.~\ref{sec:methodology}).

We want a simple, post‑hoc transform that (i) removes directions carrying \emph{identity} information and (ii) leaves the complementary directions, those most useful for non‑biometric tasks, intact.
Given labeled embeddings $z_i^j$ with per‑identity means
\begin{equation}
    \mu_i=\tfrac{1}{n}\sum_{j=1}^n z_i^j, \qquad
    \mu_C=\tfrac{1}{m}\sum_{i=1}^m \mu_i,
\end{equation}
form centered means $\Delta\mu_i=\mu_i-\mu_C$, aiming to average out image-specific variation (pose, lighting, background) while retaining between-identity differences. Next, we construct the mean matrix
$M=[\Delta\mu_1,\ldots,\Delta\mu_m]\in\mathbb{R}^{d\times m}$, whose columns' span define our estimate of $C$'s \emph{identity subspace}.
Take a thin SVD $M=U\Sigma V^\top$ and retain the top-$r$ left singular vectors $U_r\in\mathbb{R}^{d\times r}$. When $r = rank(M)$, the column space of $M$ is exactly the span of $U_r$, making $U_r$ an orthonormal basis for the column space of $M$.
Thus, we can define the orthogonal projector and sanitized feature
\begin{equation}
    P = I - U_rU_r^\top, \qquad
    \tilde{z} = \frac{P\,z}{\|P\,z\|_2}.
\end{equation}
This one‑shot, moment‑based construction depends only on class means (no covariance inversion), is numerically stable, runs in $\mathcal{O}(d\,m^2)$, and produces a fixed $P$ that we can export to any retrieval pipeline.

In the homoscedastic limit, whitening by $\Sigma_w^{-1/2}$ before taking the SVD is the textbook Fisher solution.
In practice we adopt the \emph{means‑only} variant above (no $\Sigma_w^{-1/2}$) for numerical robustness and speed; it targets the same between‑class structure and empirically suffices to collapse open‑set linear identity access at low FAR.
We select rank $r$ on held‑out identities to meet a privacy target while preserving a large complementary subspace (Sec.~\ref{sec:experiments}).

Because our threat model includes strong \emph{linear} verifiers calibrated at low FAR (Sec.~\ref{sec:methodology}), removing the between‑class mean subspace directly undermines the attacker’s decision margin: scores that relied on $w^\top(\mu_i-\mu_j)$ no longer separate unseen identities.
This yields a simple \emph{linear‑leakage certificate}: after ISP, any linear probe trained on disjoint identities operates near chance at deployment thresholds, while most non‑biometric structure is preserved in the complementary subspace.

In practice, ISP is fit \emph{offline} on a privacy-approved set of labeled identities via a single SVD; the resulting projector $P$ is a fixed $d \!\times\! d$ matrix that is exported once and applied at inference time as a single matrix multiplication, adding sub-millisecond latency to the query path. Memory and compute scale with the number of identities used during fitting, not with deployment traffic. Our cross-dataset transfer experiments (Sec.~\ref{sec:experiments}) show that a projector trained on one corpus generalizes to another with minimal degradation, suggesting that $P$ captures dataset-agnostic identity structure and need not be recomputed unless the data distribution or privacy target shifts materially.
\bigskip

\section{Experiments and Results}
\subsection{Few-shot probing}
\label{sec:experiments}

We study how much \emph{identity} is accessible in frozen embeddings at biometric operating points and how much a simple subspace removal (ISP; Sec.~\ref{sec:ISP}) suppresses that access without retraining the encoder. Throughout, we use the open‑set, low‑FAR verification protocol from Sec.~\ref{sec:probing}, and we report \emph{TAR@FAR} for unseen identities with thresholds fixed on disjoint validation identities.

To control pair counts and class balance, we curate \textbf{CelebA‑20} and \textbf{VGGFace2-20}, balanced identity subsets of CelebA and VGGFace2 respectively with exactly $n{=}20$ images per identity, aligned crops, and identity‑disjoint splits (320/80/80 train/val/test identities). We evaluate at $k\!\in\!\{1,4,16\}$ to probe attacker supervision regimes. Full pair counts, FAR denominators, thresholds, and seeds are logged in Appendix~\ref{app:probing}.

\begin{table*}[t]
    \centering
    \caption{%
        \textbf{Ridge Open-Set TAR@FAR=$10^{-4}$ (\%) Across $k$}
        TAR@FAR=$10^{-4}$ for each model over varied k, dataset, and projection method. \textbf{ISP-W} refers to the ISP projection being calculated on the same dataset, and \textbf{ISP-X} refers to the ISP projection being calculated from the other dataset. Results are averaged over 5 different seeds. FR models are \textbf{bolded}; ISP is not applied to FR encoders (— indicates not applicable).
    }
    \label{tab:bli_comprehensive}
    \scriptsize
    \begin{tabular}{@{}llccc|ccc|ccc@{}}
        \toprule
                         &                  & \multicolumn{3}{c|}{$\mathbf{k=1}$} & \multicolumn{3}{c|}{$\mathbf{k=4}$} & \multicolumn{3}{c}{$\mathbf{k=16}$}                                                                                                   \\
        \cmidrule(lr){3-5} \cmidrule(lr){6-8} \cmidrule(lr){9-11}
        \textbf{Dataset} & \textbf{Model}   & \textbf{RAW}                        & \textbf{ISP-W}                      & \textbf{ISP-X}                      & \textbf{RAW} & \textbf{ISP-W} & \textbf{ISP-X} & \textbf{RAW} & \textbf{ISP-W} & \textbf{ISP-X} \\
        \midrule
        \multirow{6}{*}{\rotatebox{90}{\textit{CelebA-20}}}
                         & DINOv2           & 4.5\%                               & 3.5\%                               & 3.7\%                               & 5.5\%        & 3.9\%          & 3.9\%          & 5.7\%        & 4.4\%          & 4.2\%          \\
                         & DINOv3           & 4.5\%                               & 2.1\%                               & 1.9\%                               & 6.7\%        & 2.4\%          & 2.2\%          & 6.8\%        & 2.8\%          & 2.5\%          \\
                         & CLIP             & 16.4\%                              & 11.9\%                              & 10.9\%                              & 19.7\%       & 15.4\%         & 10.6\%         & 19.8\%       & 13.0\%         & 10.2\%         \\
                         & SSCD             & 6.6\%                               & 3.6\%                               & 3.4\%                               & 8.4\%        & 3.9\%          & 3.9\%          & 9.8\%        & 4.4\%          & 4.5\%          \\
                         & \textbf{ArcFace} & 93.7\%                              & —                                   & —                                   & 94.0\%       & —              & —              & 94.0\%       & —              & —              \\
                         & \textbf{AdaFace} & 93.6\%                              & —                                   & —                                   & 94.0\%       & —              & —              & 94.2\%       & —              & —              \\
        \midrule
        \multirow{6}{*}{\rotatebox{90}{\textit{VGGFace2-20}}}
                         & DINOv2           & 5.0\%                               & 0.0\%                               & 3.7\%                               & 1.5\%        & 1.1\%          & 2.0\%          & 1.6\%        & 1.5\%          & 2.7\%          \\
                         & DINOv3           & 9.6\%                               & 0.4\%                               & 0.4\%                               & 3.1\%        & 0.9\%          & 1.8\%          & 5.6\%        & 0.5\%          & 2.0\%          \\
                         & CLIP             & 3.5\%                               & 0.1\%                               & 13.0\%                              & 10.0\%       & 2.7\%          & 4.6\%          & 14.1\%       & 2.3\%          & 4.6\%          \\
                         & SSCD             & 1.2\%                               & 0.1\%                               & 0.2\%                               & 1.6\%        & 0.3\%          & 0.5\%          & 4.5\%        & 0.5\%          & 0.8\%          \\
                         & \textbf{ArcFace} & 70.7\%                              & —                                   & —                                   & 80.1\%       & —              & —              & 79.6\%       & —              & —              \\
                         & \textbf{AdaFace} & 31.5\%                              & —                                   & —                                   & 65.4\%       & —              & —              & 66.1\%       & —              & —              \\
        \bottomrule
    \end{tabular}
    \vspace{0.3em}
    \footnotesize
\end{table*}

For each (dataset, model) we fit ISP once on the \emph{train} identities only. The choice of $r$ presents a trade-off---larger $r$ removes more identity energy but risks discarding task-useful variance. We sweep $r$ on a validation split disjoint from test identities and pick the smallest $r$ that meets a target privacy reduction (e.g.\ TAR@FAR$=10^{-4} \leq 5\%$). We use $r{=}256$ for DINOv2 and $r{=}192$ for DINOv3/CLIP/SSCD. The resulting $P$ is fixed and applied at test time without further tuning. To assess portability, we evaluate a $2\!\times\!2$ \emph{cross‑dataset transfer}: train $P$ on CelebA‑20 and apply it to VGGFace2 (and vice-versa), alongside the within‑dataset baselines.

Under this attacker‑calibrated protocol we observe multiple patterns. First, in raw embeddings, open‑set linear TAR at low FAR tends to be modest for DINO/SSCD and somewhat higher for CLIP, while FR controls, by design, are far higher, suggesting that strong open‑set linear probes already struggle to extract identity from non-FR embeddings. Second of all, after ISP, open‑set Ridge TAR for non-FR encoders drops to low single digits on both datasets, typically close to chance overlap. Next, the CelebA‑trained projector usually provides similar protection on VGGFace2 and vice‑versa, indicating that the dominant identity subspace we estimate is not purely dataset‑specific; we see small asymmetries by model, but overall transfer remains strong. Finally, we notice that trends are stable from $k{=}1$ through $k{=}16$: ISP remains protective and relative model ordering changes only slightly, which is encouraging for few‑shot and many‑shot attacker regimes alike (Table~\ref{tab:bli_comprehensive}).

The transfer we see is consistent with a low‑rank set of directions that carry most of the linearly-accessible identity variance. Our principal‑angle analysis confirms this: identity subspaces fitted on disjoint datasets exhibit high alignment (cosine of leading principal angles ${>}0.99$ for all models), supporting the view that identity concentrates in a compact, portable subspace. In practice, an operator can train \emph{one} ISP once (on a privacy‑approved identity set), then ship the fixed projector $P$ to new corpora with minimal re‑tuning (Table~\ref{tab:bli_comprehensive}). Full principal‑angle results, and sensitivity analyses for rank $r$ and the number of training identities are provided in Appendix~\ref{app:isp-sensitivity}.


\begin{figure}[t]
    \centering
    \includegraphics[width=\linewidth]{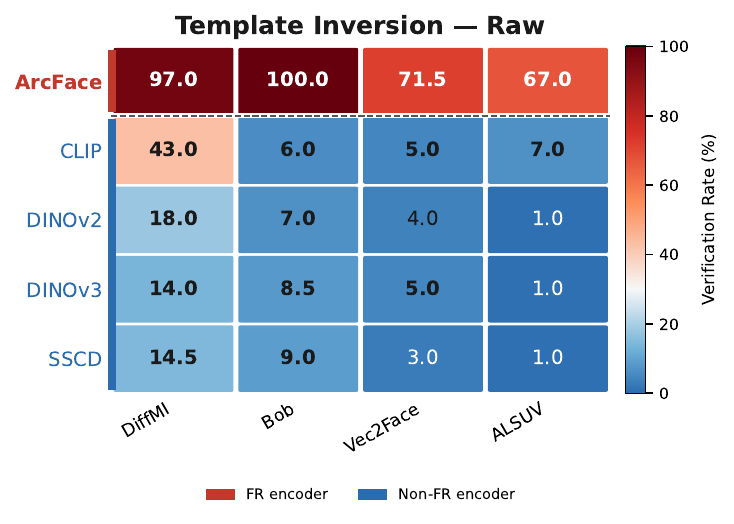}
    \caption{%
        \textbf{Template Inversion Verification Rate (\%).}
        Heatmap of verification rates across four attack families (DiffMI, Bob, Vec2Face, ALSUV) and five encoders.
        The FR encoder (ArcFace, top row) achieves consistently high rates (67--100\%), while non-FR encoders (CLIP, DINOv2, DINOv3, SSCD) remain near zero across all methods.
        All rates averaged over ArcFace ($\tau{=}0.1051$) and AdaFace ($\tau{=}0.1111$) verifiers.
    }
    \label{fig:inversion_verify}
\end{figure}

\subsection{Non-Linear Robustness}
\label{sec:nonlinear}

ISP is designed to certify against \emph{linear} attackers; a natural concern is whether identity information persists in directions that a non-linear probe can exploit. To test this, we train a projection-only MLP verifier (two hidden layers, ReLU activations) under the same open-set, identity-disjoint protocol as the Ridge probe above. The MLP is trained on $A_{\text{train}}$ with threshold and hyperparameters frozen on $A_{\text{val}}$ before evaluation on $A_{\text{test}}$, ensuring a fair comparison.

Results are shown in Tables~\ref{tab:mlp_verifier_full} and~\ref{tab:mlp_verifier_vgg} (Appendix~\ref{app:mlp}) for CelebA-20 and VGGFace2-20 respectively. After ISP, the MLP TAR drops to near zero across both datasets. FR baselines remain largely unaffected, as expected.

These results suggest that while ISP's formal guarantees are linear, the identity subspace it removes can also carry signal available to moderate non-linear attackers. We emphasize that this does not constitute a non-linear certificate, nor is it substantive evidence to confirm the phenomenon: sufficiently expressive models with enough data could in principle recover residual cues.

\subsection{Utility Retention}
\label{sec:utility}

\begin{figure*}[t]
    \centering
    \includegraphics[width=\textwidth]{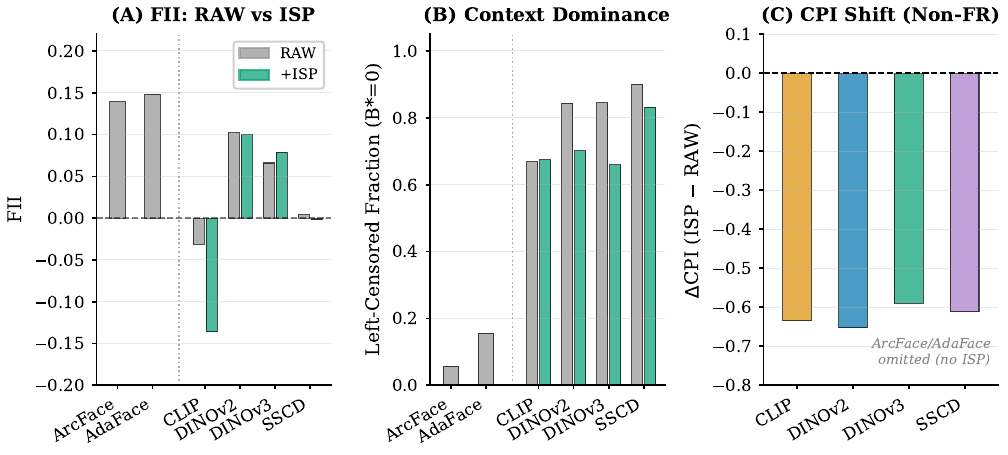}
    \caption{%
        \textbf{FCR Attribution: RAW vs ISP Overlay.}
        (A) FII comparison: ISP reduces face-driven similarity.
        (B) Censoring comparison: context dominance maintained.
        (C) CPI delta: dramatic shift from context-only (1.0) to balanced (0.35-0.41).
        All images were cropped to a FCR of 0.33, with $\pm$2\% equal-area tolerance.
    }
    \label{fig:fcr_ISP}
\end{figure*}

We fit each projector once on \textbf{VGGFace2-20} (320/80/80 train/val/test identities; RetinaFace crops)
and apply it unchanged to downstream utility tasks.
\textbf{ISP} is computed via a one-shot SVD on class means with rank $r$ chosen by validation variance coverage
(DINOv2: $r{=}256$, others: $r{=}192$). We also evaluate and compare utility preservation against another projection method, \textbf{LEACE}\cite{Belrose2023LEACE}.
\textbf{LEACE} uses a closed-form least-squares erasure with
$\lambda\!\in\!\{10^{-6},10^{-5},10^{-4},10^{-3},10^{-2}\}$ selected on validation to minimize leakage.
We then measure downstream utility on ImageNet with frozen $k$-NN and linear-probe classification,
and on DISC2021 copy detection for SSCD (see Table~\ref{tab:utility_disc_raw} in Appendix~\ref{app:disc}).

ImageNet and DISC2021 accuracies are normalized to the unprojected baseline for each model
(\emph{100 = raw performance}).
Values below 100 therefore indicate a relative reduction in utility after projection.

We find that both ISP and LEACE preserve nearly all downstream utility, aside from SSCD, which is task specific and not intended to be used for semantic classification.
Across models, ImageNet classification accuracy remains close to 100\% of baseline,
and DISC2021 copy-detection performance drops by only a few percent.
Small differences between ISP and LEACE are within experimental noise.
Overall, we observe \emph{minimal utility loss} from applying privacy-oriented projections
trained on facial embeddings to unrelated visual tasks.

\begin{table}[t]
    \centering
    \caption{\textbf{ImageNet utility (normalized; 100 = raw baseline).}
        Top-1 accuracy (percentage of raw, unprojected baseline) for $k$-NN and linear probe after projection.}
    \label{tab:utility_imagenet_norm}
    \small
    \setlength{\tabcolsep}{5pt}
    \begin{tabular}{lcc|cc}
        \toprule
        \multirow{2}{*}{\textbf{Model}}           &
        \multicolumn{2}{c}{\textbf{$k$-NN Top-1}} &
        \multicolumn{2}{c}{\textbf{Linear Probe Top-1}}                                                           \\
        \cmidrule(lr){2-3}\cmidrule(lr){4-5}
                                                  & \textbf{ISP} & \textbf{LEACE} & \textbf{ISP} & \textbf{LEACE} \\
        \midrule
        DINOv2                                    & 100.1\%      & 100.0\%        & 99.2\%       & 99.3\%         \\
        DINOv3                                    & 97.3\%       & 97.4\%         & 93.5\%       & 93.4\%         \\
        CLIP                                      & 98.3\%       & 98.5\%         & 100.7\%      & 105.6\%        \\
        SSCD                                      & 85.4\%       & 85.7\%         & 83.3\%       & 82.6\%         \\
        \bottomrule
    \end{tabular}
\end{table}

\subsection{Context Attribution}
\label{sec:r4}
We FCR-normalize faces and compute three complementary diagnostics on tightly cropped portrait pairs: \emph{FII} (equal-area face vs.\ matched background occlusions), \emph{CPI} (face-blur series comparing same-person/different-context vs.\ different-person/same-context), and a \emph{stress test} $B^*$ (monotone background revelation until context overtakes identity, with censoring at the boundaries). All perturbations are strength-matched and equal-area; implementation details and mask sanity checks are in Appendix~\ref{app:fca}.

We find that identity-focused FR baselines are face-dominant (occluding the face hurts most, and they resist background revelation), whereas non-FR encoders inside tight crops are context-dominant under the stress test (background can outweigh face when forced). After ISP, the most conspicuous change is in \emph{context preference}: models that previously over-rely on same-context signals rebalance toward same-person preference, while the extreme stress test remains extreme by construction. Face-occlusion sensitivity either decreases or remains low, indicating that ISP has removed face-aligned identity evidence without creating brittle behavior elsewhere (Figure~\ref{fig:fcr_ISP}). Thus, we find that these diagnostics separate FR models (face‑dominant: FII$>$0, $B^*\!\to\!1$) from non-FR encoders (context‑dominant under stress: FII$\approx$0, $B^*\!\to\!0$) on VGGFace2 under open‑set protocols, and remain stable across pairing strategies, providing a quantitative complement to prior saliency/attention methods.

\subsection{Template Inversion}
\label{sec:r5}
A single inversion method cannot adequately characterize generative identity leakage, as success depends on the generative prior, optimization strategy, and budget. We therefore evaluate four attack families, DiffMI~\cite{Wang2025} (diffusion-based), Vec2Face~\cite{Wu2025Vec2Face} (regression-based), Bob~\cite{Kim2024Scores} (score-based), and ALSUV~\cite{Jung2024ALSUV} (latent-optimization-based), under matched budgets and judge success via cross-model FR verification (Sec.~\ref{sec:inversion}).

\begin{figure}[h]
    \centering
    \includegraphics[width=\columnwidth]{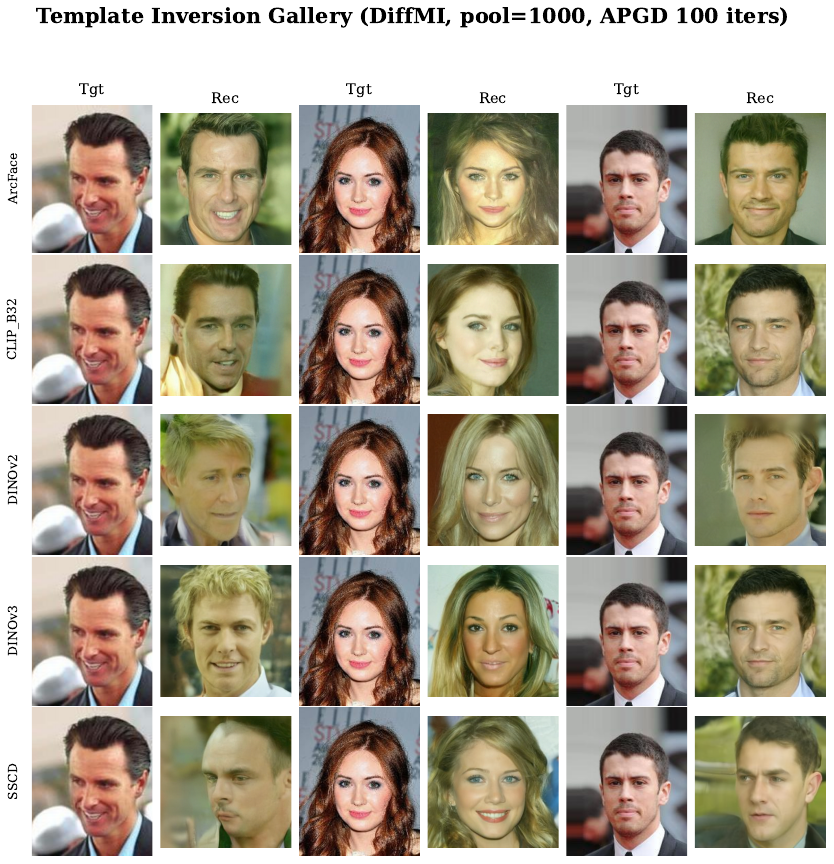}
    \caption{%
        \textbf{Template Inversion Gallery.}
        Target vs.\ DiffMI reconstruction across five encoders.
        ArcFace reconstructions verify at high rates; non-FR encoders
        do not.
        Protocol: DiffMI white-box, 1{,}000-latent pool, APGD
        100 iterations, $N{=}200$ CelebA-20 targets.
    }
    \label{fig:inversion_gallery}
\end{figure}

Figure~\ref{fig:inversion_verify} summarizes results across all methods, and Figure~\ref{fig:inversion_gallery} visualizes DiffMI results.. FR encoders invert reliably regardless of attack family: ArcFace and AdaFace achieve 67--100\% verification depending on the method, confirming that dedicated FR embeddings carry a strong, prior-navigable identity signal. In contrast, non-FR encoders (CLIP, DINOv2, DINOv3, SSCD) remain at or near zero for DiffMI and reach only low single digits for Bob, Vec2Face, and ALSUV. The gap between FR and non-FR encoders is consistent across all four attack families, indicating that the low inversion success is not an artifact of a single method's limitations.

Critically, applying ISP does not materially increase or decrease inversion success for non-FR encoders: raw and post-ISP verification rates are comparable across all methods. This is important because removing a linear subspace could, in principle, expose residual structure that a generative attacker could exploit; empirically, over the attack surfaces we explore, it does not. While ISP provides no formal guarantee against inversion (which depends on the generative prior and optimization budget), the empirical evidence suggests that the identity signal in non-FR embeddings is too weak for current inversion methods to operationalize - both before and after projection.

\section{Limitations and Future Work}
Our study is restricted to frozen encoders and evaluates only \emph{linear} susceptibility; ISP provides no guaranteed protection against stronger non-linear or generative attacks. Our identity subspace estimates rely on labeled identities and may be sensitive to demographic imbalance or domain mismatch, though our cross-dataset transfer results suggest good robustness. ISP also targets between-class mean structure only; higher-order identity cues may remain in the complementary space. Future work should evaluate privacy under black-box and gradient-free adversaries, extend ISP with iterative or non-linear variants that retain its auditability, and explore training-time integration to obtain encoders that are privacy-preserving by design. Finally, scaling our attribution and inversion audits to broader datasets and sensitive attributes beyond identity remains an important direction for deployment-ready privacy assessments.

\section{Conclusion}
We presented a comprehensive study of identity leakage in visual embeddings and introduced Identity Sanitization Projection (ISP) as an effective mitigation. Our proposed metrics (Open-set TAR@low-FAR, inversion success, FCR, CPI, $B^*$) allowed us to quantify leakage and show that models like CLIP and DINOv2 expose minimal accessibility of identity. ISP leverages linear subspace removal to further dampen this information, providing substantial privacy gains with minimal impact on the embeddings' usefulness for other tasks. We provided theoretical justification and practical evidence for ISP's effectiveness, and through ablation and attribution studies, we showed that identity evidence in non-FR encoders concentrates in a compact, transferable linear subspace, enabling a strong linear privacy-utility trade‑off via ISP.

This work takes a step toward making powerful vision models more privacy-friendly, which is crucial for real-world deployment. By open-sourcing our evaluation toolkit and encouraging others to adopt privacy metrics, we hope to spur development of even more advanced mitigation techniques. Ultimately, we envision non-FR embeddings that can retain their general utility while offering guarantees that using them will not compromise individuals' privacy.

\section*{Acknowledgements}
The authors would like to thank Jinxing Li, Hongkai Pan, James Chang, Nicholas Ren, Amy Du, and Saatvik Billa for assistance and feedback.


\clearpage
\setcounter{page}{1}

\appendix

\section{Few-Shot Verification: Full Protocol}
\label{app:probing}

This section provides complete implementation-ready details for the
open-set verification experiments in Sec.~4.1, including dataset
curation, pair construction, threshold calibration, hyperparameter
selection, and resampling.

\subsection{Dataset Curation}
\label{app:datasets}

We curate two balanced identity subsets with strict identity separation:

\paragraph{CelebA-20.} 6,348 identities with exactly 20 images each
(126,960 total images). Open-set split: 320 train / 80 val / 80 test
identities (1,600 test images).

\paragraph{VGGFace2-20.} 480 identities with exactly 20 images each
(9,600 total images). Open-set split: 320 train / 80 val / 80 test
identities (1,600 test images).

All splits are identity-disjoint:
$\mathcal{A}_{\text{train}} \cap \mathcal{A}_{\text{val}} \cap
    \mathcal{A}_{\text{test}} = \varnothing$, ensuring no identity appears
in multiple splits.

The full CelebA and VGGFace2 datasets contain variable numbers of
images per identity and heterogeneous crops (tight face crops, wider
scenes, varying resolutions). Our balanced subsets standardize these
nuisance factors so that FAR calibration and per-identity pair sampling
are comparable across probes, models, and projection methods. This is
important because unbalanced identity counts would distort the impostor
distribution and make FAR estimates unreliable across experimental
conditions.

\subsection{Pair Construction}
\label{app:pairs}

For each identity $i$ in a given split, we form mated (same-identity)
and impostor (different-identity) pairs exclusively from the query set
$Q_i$; the support set $S_i$ is used only to train the probe and never
appears in evaluation pairs:
\begin{align}
    \mathcal{P}_{\text{mated}}
     & = \{(z_a, z_b) : a, b \in Q_i,\; a \neq b\},
    \label{eq:mated}                                         \\
    \mathcal{P}_{\text{impostor}}
     & = \{(z_a, z_b) : a \in Q_i,\;
    b \in Q_j,\; i \neq j\}.
    \label{eq:impostor}
\end{align}
Mated pairs are exhaustive within an identity: for an identity with $n$
embeddings split into $|S_i| = k$ support and $|Q_i| = n - k$ query
samples, all $\binom{n-k}{2}$ within-identity query pairs are included.

Impostor pairs are subsampled with \emph{identity-balanced quotas}: for
each identity $i$, we sample a fixed number of cross-identity
comparisons per other identity $j$, ensuring that no single identity
pair dominates the impostor distribution. This stabilizes the FAR
denominator across different values of $k$ and across resampling seeds.

\subsection{Pair Counts and FAR Precision}
\label{app:pair-counts}

FAR precision is $1/|\mathcal{P}_{\text{impostor}}|$, representing the
smallest measurable FAR increment. Table~\ref{tab:pair_counts} reports
exact counts for each dataset.

\begin{table}[h]
    \centering
    \caption{Test set pair counts. Min.\ FAR shows the minimum
    measurable FAR ($1/n_{\text{impostor}}$).}
    \label{tab:pair_counts}
    \small
    \begin{tabular}{lcccc}
        \toprule
        \textbf{Dataset} & \textbf{Test IDs} & \textbf{Mated}
                         & \textbf{Impostor} & \textbf{Min.\ FAR}                                           \\
        \midrule
        CelebA-20        & 80                & 15,200             & ${\sim}6{,}400$  & $1.6 \times 10^{-4}$ \\
        VGGFace2-20      & 80                & 15,200             & ${\sim}16{,}000$ & $6.3 \times 10^{-5}$ \\
        LFW-20           & 13                & 78                 & 1,326            & $7.5 \times 10^{-4}$ \\
        \bottomrule
    \end{tabular}
    \vspace{0.3em}
    \footnotesize
    CelebA-20 and VGGFace2-20 evaluated at TAR@FAR$=10^{-4}$.
    LFW-20 uses partial AUC at FAR$\in[0, 10^{-3}]$ due to
    insufficient impostor pairs.
\end{table}

\subsection{Threshold Calibration}
\label{app:threshold}

We calibrate a verification threshold $\tau$ on validation identities
$\mathcal{A}_{\text{val}}$ only. The procedure is as follows:

\begin{enumerate}[nosep]
    \item Compute all mated and impostor similarity scores on
          $\mathcal{A}_{\text{val}}$.
    \item When the impostor count permits
          ($|\mathcal{P}_{\text{impostor}}^{\text{val}}| \geq 10{,}000$),
          select $\tau$ such that
          $\mathrm{FAR}(\tau) \approx 10^{-4}$.
    \item When the impostor count is insufficient for stable
          $10^{-4}$ estimation (e.g., LFW-20), fall back to a
          fixed-head partial AUC criterion at
          FAR$\in[0, 10^{-3}]$ and still return a single $\tau$.
    \item A \emph{separate} $\tau$ is calibrated for each (dataset,
          encoder, projection method, $k$) combination.
    \item Once set on validation identities, $\tau$ is
          \emph{held fixed} for all test resamplings---the five seeds
          that reshuffle $S_i / Q_i$ per identity do not each get
          their own threshold.
\end{enumerate}

\noindent This ensures that no information from test identities
influences the operating point. We report TAR on test along with the
achieved FAR (as both a rate and raw count of false accepts over total
impostor comparisons) so that readers can verify the operating point is
meaningful.

\subsection{Hyperparameter Selection}
\label{app:hyperparams}

We train a ridge regression verifier on support embeddings from
$\mathcal{A}_{\text{train}}$ and evaluate TAR@FAR on
$\mathcal{A}_{\text{test}}$. We sweep $L_2$ regularization strength
$\alpha \in \{10^{-3}, 10^{-2}, 10^{-1}, 1, 10\}$ on
$\mathcal{A}_{\text{val}}$ and select the setting that maximizes
verification performance under the target operating point
($\mathrm{TAR}@\mathrm{FAR} \approx 10^{-4}$). The chosen $\alpha$
and threshold $\tau$ are frozen before any evaluation on
$\mathcal{A}_{\text{test}}$.

\subsection{$k$-Shot Sampling and Resampling}
\label{app:resampling}

For $k$-shot evaluation where $k \in \{1, 4, 16\}$:

\begin{enumerate}[nosep]
    \item For each identity $i$, randomly split $B_i$ into query
          set $Q_i$ and support set $S_i$ with $|S_i| = k$.
    \item Train the probe (Ridge or MLP) on support embeddings from
          training identities only.
    \item Calibrate $\tau$ on validation identities (as above).
    \item Evaluate on test identities using the fixed $\tau$.
    \item Repeat with 5 random seeds for $S_i / Q_i$ assignment;
          report mean TAR and 95\% identity-aware confidence
          intervals.
\end{enumerate}

\noindent ``Identity-aware'' refers to the fact that variance is
computed across seeds (which reshuffle per-identity splits) rather than
across individual pairs, avoiding pseudo-replication from correlated
pairs within the same identity.

\subsection{LFW-20 Sanity Check}
\label{app:lfw}

LFW-20 has insufficient impostor pairs (1,326) for reliable
TAR@FAR$=10^{-4}$ calibration (minimum measurable
FAR$=7.5 \times 10^{-4}$). We include it as a sanity check using
partial AUC at FAR$\in[0, 10^{-3}]$ and do not draw conclusions from
it in the main paper.

\begin{table}[h]
    \centering
    \caption{LFW-20 Ridge Open-Set: partial AUC (\%) at
        FAR$\in[0, 10^{-3}]$ (13 test IDs). ISP-W is not applied to FR
        models.}
    \label{tab:lfw}
    \small
    \begin{tabular}{l ccc ccc}
        \toprule
                       & \multicolumn{3}{c}{\textbf{RAW}}
                       & \multicolumn{3}{c}{\textbf{ISP-W}}                                      \\
        \cmidrule(lr){2-4} \cmidrule(lr){5-7}
        \textbf{Model} & $k$=1                              & $k$=4 & $k$=16
                       & $k$=1                              & $k$=4 & $k$=16                     \\
        \midrule
        DINOv2         & 19.5                               & 6.4   & 9.5    & 0.0  & 0.0  & 0.0 \\
        DINOv3         & 11.5                               & 0.8   & 6.7    & 0.0  & 10.3 & 2.1 \\
        CLIP           & 55.6                               & 40.8  & 9.0    & 10.8 & 15.4 & 0.0 \\
        SSCD           & 2.3                                & 14.1  & 0.0    & 2.1  & 0.0  & 0.5 \\
        ArcFace        & 87.4                               & 82.1  & 73.9   & ---  & ---  & --- \\
        AdaFace        & 85.4                               & 94.4  & 67.7   & ---  & ---  & --- \\
        \bottomrule
    \end{tabular}
\end{table}

\clearpage

\section{ISP Sensitivity and Cross-Dataset Transfer}
\label{app:isp-sensitivity}

This section provides the full sensitivity analyses and transfer
experiments justifying our ISP rank selection and fitting procedure.
All experiments use the within-dataset (ISP-W) protocol, Ridge attacker,
TAR@FAR$=10^{-4}$, and 3 evaluation seeds unless otherwise noted.

\subsection{Cross-Dataset Transfer and Principal Angles}
\label{app:cross-dataset}

\paragraph{Transfer Protocol.}
To evaluate whether ISP generalizes across datasets:
\begin{enumerate}[nosep]
    \item Fit ISP on dataset A (e.g., CelebA-20, 320 train identities).
    \item Apply $P_A$ to embeddings from dataset B (e.g., VGGFace2-20).
    \item Calibrate threshold on dataset B validation set.
    \item Evaluate on dataset B test set.
\end{enumerate}

We evaluate a full $2 \times 2$ transfer matrix:
\begin{itemize}[nosep]
    \item Within-dataset: $P_A \to A$, $P_B \to B$
    \item Cross-dataset: $P_A \to B$, $P_B \to A$
\end{itemize}

\paragraph{Principal Angle Analysis.}
To quantify subspace alignment between projections fitted on different
datasets, we compute principal angles between the identity subspaces
$U_A$ and $U_B$:
\begin{equation}
    \cos(\theta_i) = \sigma_i(U_A^\top U_B)
\end{equation}
where $\sigma_i$ are the singular values of $U_A^\top U_B$, giving the
cosines of principal angles.

\noindent\textbf{Cross-dataset alignment} (same model, different datasets):
\begin{itemize}[nosep]
    \item DINOv2: $\max \cos(\theta) = 0.9998$
    \item DINOv3: $\max \cos(\theta) = 1.0000$
    \item CLIP: $\max \cos(\theta) = 0.9998$
    \item SSCD: $\max \cos(\theta) = 0.9977$
\end{itemize}

Near-perfect alignment ($\cos \theta \approx 1.0$) indicates identity
subspaces are \emph{universal} across datasets, not dataset-specific
artifacts.

\subsection{Rank Sweep}
\label{app:rank-sweep}

Table~\ref{tab:rank_sweep} shows TAR as $r$ increases from 0 (raw, no
projection) through 64, 96, 128, and 192 on both CelebA-20 and
VGGFace2-20 with $m{=}320$ training identities.
Protection generally improves with rank; CLIP requires higher $r$ due
to its larger embedding dimension and broader identity subspace.
For VGGFace2-20, DINOv2/v3/SSCD reach 0.0\% TAR already at moderate
ranks, indicating near-complete suppression of linear identity access.

\begin{table}[h]
    \centering
    \small
    \setlength{\tabcolsep}{5pt}
    \caption{%
    \textbf{Rank sweep.} TAR@FAR$=10^{-4}$ (\%) under ISP-W
    (Ridge, $m{=}320$, $k{=}16$). $r{=}0$ is the raw (unprojected)
    baseline.
    }
    \label{tab:rank_sweep}
    \begin{tabular}{@{}ll ccccc@{}}
        \toprule
        \textbf{Dataset} & \textbf{Model} & $r{=}0$ & $r{=}64$ & $r{=}96$     & $r{=}128$    & $r{=}192$    \\
        \midrule
        \multirow{4}{*}{CelebA-20}
                         & DINOv2         & 4.1     & 3.7      & 2.9          & \textbf{1.3} & 1.5          \\
                         & DINOv3         & 6.2     & 2.3      & 1.4          & \textbf{1.0} & 1.5          \\
                         & SSCD           & 2.9     & 1.4      & 1.5          & \textbf{0.6} & 1.0          \\
                         & CLIP           & 7.3     & 6.6      & 6.7          & 5.7          & \textbf{3.5} \\
        \midrule
        \multirow{4}{*}{VGGFace2-20}
                         & DINOv2         & 3.1     & 3.5      & 3.1          & 2.7          & \textbf{2.2} \\
                         & DINOv3         & 2.9     & 2.5      & \textbf{1.0} & 2.2          & 1.5          \\
                         & SSCD           & 1.3     & 2.0      & 2.1          & 1.8          & \textbf{1.0} \\
                         & CLIP           & 7.1     & 4.7      & \textbf{2.9} & 4.2          & 2.3          \\
        \bottomrule
    \end{tabular}
\end{table}

\subsection{Identity-Count Sweep}
\label{app:identity-sweep}

Table~\ref{tab:identity_sweep} shows how TAR changes as the number of
identities $m$ used to fit the ISP projector scales from 320 to 2000,
using the ISP-Large protocol (fit on a held-out pool disjoint from test
identities, evaluated on CelebA-20 at $r{=}192$).
Protection does not degrade with more identities; CLIP shows the largest
improvement at scale ($m{=}2000$: 1.5\% vs.\ $m{=}320$: 3.5\%),
suggesting its identity subspace benefits from a larger fit pool.
DINOv3 and SSCD already reach sub-1\% TAR at $m{=}320$ and remain stable.

\begin{table}[h]
    \centering
    \small
    \setlength{\tabcolsep}{6pt}
    \caption{%
    \textbf{Identity-count sweep.} TAR@FAR$=10^{-4}$ (\%) as the
    ISP fit pool $m$ scales (ISP-Large, CelebA-20, Ridge,
    $r{=}192$, $k{=}16$). Fit identities are disjoint from test
    identities.
    }
    \label{tab:identity_sweep}
    \begin{tabular}{@{}l cccc@{}}
        \toprule
        \textbf{Model} & $m{=}320$    & $m{=}640$ & $m{=}1280$ & $m{=}2000$   \\
        \midrule
        DINOv2         & 1.5          & 1.9       & 2.9        & 2.7          \\
        DINOv3         & 1.5          & 5.4       & 4.0        & 3.9          \\
        SSCD           & 1.0          & 1.3       & 1.7        & 1.7          \\
        CLIP           & \textbf{3.5} & 6.7       & 6.7        & \textbf{1.5} \\
        \bottomrule
    \end{tabular}
\end{table}

\clearpage

\section{Template Inversion: Protocols and Extended Results}
\label{app:inversion}

This section provides per-attack protocol details, the shared
cross-model verification criterion, and extended quantitative results.

\subsection{Shared Evaluation Protocol}
\label{app:inversion-eval}

All template inversion attacks are evaluated under the same
cross-model verification criterion. Given a reconstructed face
$\hat{x}$ and true target $x_{\text{tgt}}$, inversion succeeds if

\begin{equation}
    \frac{f_{\text{FR}}(\hat{x})^\top f_{\text{FR}}(x_{\text{tgt}})}
    {\|f_{\text{FR}}(\hat{x})\|_2\,\|f_{\text{FR}}(x_{\text{tgt}})\|_2}
    \;\geq\; \tau_F,
\end{equation}

where $f_{\text{FR}}$ is a \emph{disjoint} FR encoder (different from
the one whose embedding is being inverted). We evaluate with two FR
verifiers:

\begin{itemize}[nosep]
    \item \textbf{ArcFace}: $\tau_F = 0.1051$ (minimum EER on validation).
    \item \textbf{AdaFace}: $\tau_F = 0.1111$ (minimum EER on validation).
\end{itemize}

\noindent Verification rate is $\text{\# accepted} / \text{\# targets} \times 100\%$;
all values in the main paper and tables below are averaged over both verifiers.
95\% binomial confidence intervals are computed via the Wilson score method.

\subsection{DiffMI (Wang et al., 2025)}
\label{app:diffmi}

DiffMI is a training-free white-box attack that uses an unconditional
face diffusion prior to reconstruct faces from target embeddings alone.

\paragraph{Candidate pool generation.}
\begin{itemize}[nosep]
    \item Pre-generate $V = 1{,}000$ latent codes
          $\ell \sim \mathcal{N}(0, I)$ with shape $[3, 256, 256]$.
    \item Decode via DDIM (20-step denoising) to produce
          $256 \times 256$ face images.
    \item Validate each latent: K$^2$ normality test ($p \geq 0.999$)
          and MTCNN face detection (confidence $\geq 0.999$).
          Acceptance rate $\approx 10\%$; validated pool cached for reuse.
\end{itemize}

\paragraph{Top-$N$ selection.}
For each target embedding $z_{\text{tgt}}$, compute cosine similarity
with all pool images and select the $N = 3$ candidates with highest
similarity as warm-start initialization.

\paragraph{Adversarial refinement (APGD).}
\begin{itemize}[nosep]
    \item Objective: maximize
          $\cos\bigl(f(G(\ell)),\, z_{\text{tgt}}\bigr)$
          where $G$ is the DDPM decoder and $f$ is the target encoder.
    \item $L_2$-bounded perturbations: $\|\Delta\ell\|_2 \leq \varepsilon = 35$.
    \item Budget: 100 iterations per target.
    \item Early stopping: halt if cosine $\geq \tau_C = 0.99$.
\end{itemize}

\paragraph{Diffusion prior.}
Unconditional DDPM trained on CelebA-HQ; DDIM decoder with 20
denoising steps; output $256 \times 256$ RGB.

\paragraph{Models and targets.}
\begin{itemize}[nosep]
    \item \textbf{ArcFace} (512-dim), \textbf{CLIP ViT-B/32} (512-dim),
          \textbf{DINOv2-base} (768-dim), \textbf{DINOv3-small} (384-dim),
          \textbf{SSCD-ResNet50} (512-dim).
    \item $N = 200$ targets from CelebA-20 test set.
\end{itemize}

\subsection{Bob (Kim et al., 2024)}
\label{app:bob}

Bob~\cite{Kim2024Scores} algebraically inverts face recognition
embeddings without iterative optimization, using only cosine similarity
scores against a small set of probe images.

\paragraph{Method overview.}
\begin{enumerate}[nosep]
    \item Build a local ArcFace embedding space model (iresnet50,
          512-dim).
    \item Pre-generate 99 $\delta$-orthogonal face images (orthogonal
          face set; OFS).
    \item For each target embedding $z_{\text{tgt}}$: compute 99
          cosine similarity scores against OFS images, solve
          $z \approx A^{\dagger} s$ via pseudo-inverse where $s$ is
          the score vector and $A$ encodes the OFS geometry, then
          decode $\hat{x} = \text{NbNet}(z)$ via the NbNet inverse
          model.
\end{enumerate}

\paragraph{Key parameters.}
\begin{itemize}[nosep]
    \item \textbf{Local model}: ArcFace iresnet50 (512-dim).
    \item \textbf{Inverse model}: NbNet (512-dim embedding $\to$ $128\!\times\!128$ face).
    \item \textbf{OFS size}: 99 probe images.
    \item \textbf{Budget}: 99 embedding queries per target (closed-form; no iterations).
\end{itemize}

\paragraph{Non-FR encoder adaptations.}
The same OFS and NbNet decoder are used across all encoders.
Encoder-specific preprocessing wrappers (resize, center-crop, or
MTCNN alignment) ensure inputs are compatible.

\paragraph{Targets.}
$N = 200$ targets from CelebA-20 test set.

\subsection{Vec2Face (Wu et al., 2025)}
\label{app:vec2face}

Vec2Face~\cite{Wu2025Vec2Face} uses a ViT-based conditional face
generator trained to produce faces whose embeddings match a target
vector.

\paragraph{Protocol for FR encoders.}
Use the pretrained Vec2Face model (trained on ArcFace embedding space)
directly for ArcFace targets.

\paragraph{Protocol for non-FR encoders.}
Train a separate Vec2Face model per embedding type:
\begin{itemize}[nosep]
    \item Embedding dimensions: CLIP (512), DINOv2 (768), DINOv3 (384),
          SSCD (512).
    \item Training: 19 epochs, lr$= 0.001$ with MultiStepLR decay,
          batch size 32, cosine similarity loss.
\end{itemize}

\paragraph{Inference budget.}
100 optimization steps per target with temporal latent averaging over
the last 70 steps; top-$K{=}10$ candidate selection.

\paragraph{Targets.}
$N = 200$ targets (FR); $N = 200$ targets per non-FR encoder (reduced
due to per-encoder training cost).

\subsection{ALSUV (Jung et al., 2024)}
\label{app:alsuv}

ALSUV~\cite{Jung2024ALSUV} searches a StyleGAN2 W$+$ latent space via
Adam with multi-latent parallel search and unsupervised validation via
latent averaging.

\paragraph{Generator and latent space.}
StyleGAN2 FFHQ; W$+$ space (14 layers $\times$ 512 dims).

\paragraph{Optimization.}
\begin{itemize}[nosep]
    \item 10 latents per batch, optimized in parallel.
    \item Adam: $\text{lr}=0.01$, weight decay $10^{-4}$, LR drop at
          step 50 (factor $0.1$).
    \item Budget: 100 steps per target.
    \item Latent averaging: average over the last 70 steps of the
          optimization trajectory (unsupervised validation).
\end{itemize}

\paragraph{Non-FR encoder adaptations.}
Encoder adapters for CLIP, DINOv2, DINOv3, and SSCD with optional
ISP projection support are applied before the cosine similarity
objective.

\paragraph{Targets.}
$N = 100$ targets from CelebA-20 test set.

\newpage
\subsection{Extended Results and ISP Comparison}
\label{app:inversion-results}

Table~\ref{tab:inversion_isp} reports raw and post-ISP verification
rates for non-FR encoders across all attack families. ISP has no
applicable variant for FR encoders (ArcFace/AdaFace). Rates are
averaged over ArcFace and AdaFace verifiers.

\begin{table}[t]
    \centering
    \caption{%
        \textbf{Template inversion verification rates (\%), RAW vs.\ +ISP.}
        Non-FR encoders only (ISP not applicable to FR encoders).
        $N{=}200$ per encoder; ALSUV $N{=}100$.
        Values averaged over ArcFace ($\tau{=}0.1051$) and
        AdaFace ($\tau{=}0.1111$) verifiers.
    }
    \label{tab:inversion_isp}
    \small
    \setlength{\tabcolsep}{4.5pt}
    \begin{tabular}{@{}l cc cc cc@{}}
        \toprule
                                              &
        \multicolumn{2}{c}{\textbf{Bob}}      &
        \multicolumn{2}{c}{\textbf{Vec2Face}} &
        \multicolumn{2}{c}{\textbf{ALSUV}}                                          \\
        \cmidrule(lr){2-3}\cmidrule(lr){4-5}\cmidrule(lr){6-7}
        \textbf{Encoder}                      & RAW & +ISP & RAW & +ISP & RAW & +ISP \\
        \midrule
        CLIP                                  & 6.0 & 6.5  & 5.0 & 5.0  & 7.0 & 2.0  \\
        DINOv2                                & 7.0 & 8.5  & 4.0 & 5.0  & 1.0 & 2.0  \\
        DINOv3                                & 8.5 & 7.5  & 5.0 & 5.0  & 1.0 & 0.0  \\
        SSCD                                  & 9.0 & 5.5  & 3.0 & 3.0  & 1.0 & 0.0  \\
        \bottomrule
    \end{tabular}
\end{table}

\clearpage

\section{Face--Context Attribution: Complete Methods}
\label{app:fca}

This section provides extended related work, formal definitions, and
full implementation details for the three face--context attribution
diagnostics (FII, CPI, $B^*$) described in Sec.~3.2.

\subsection{Extended Related Work}
\label{app:fca-related}

Modern encoders can entangle identity evidence with non-facial context
(backgrounds, clothing, hair, scene), a phenomenon widely studied in
object recognition as ``context bias'' or ``shortcut'' reliance. Classic
analyses show CNNs tend toward local texture and background
cues~\cite{Geirhos2019,Xiao2020}, while transformer backbones propagate
long-range information via self-attention, amplifying non-local context
unless constrained~\cite{Chefer2021}. In biometrics, explainable AI work
has primarily focused on \emph{where} a face recognition (FR) model
looks---via occlusion/perturbation maps and saliency---rather than
quantifying the \emph{relative importance} of face vs.\ context.
Representative perturbation methods include meaningful
masking~\cite{Fong2017} and randomized occlusion
(RISE)~\cite{Petsiuk2018}, and recent black-box explanations tailored to
faces (e.g., MinPlus~\cite{Mery2022MinPlus}) produce stable,
interpretable saliency for FR match decisions.

Beyond visualization, several studies probe context attribution
quantitatively. In generic recognition, Adhikari~et~al.\ report
``volume attribution'' that partitions saliency mass between object and
background, finding non-trivial background reliance even for correct
predictions~\cite{Adhikari2024LostInContext}. For vision--language
encoders, CLIP has been shown to produce attention and saliency that
sometimes concentrates on background rather than foreground, motivating
architectural or inference-time adjustments such as attention surgery and
segmentation-aware debiasing~\cite{Li2023CLIPExplain,Wu2025SegDebias}.
For face embeddings specifically, open-set attribution is more
challenging: similarity scores are defined over image \emph{pairs}, so
explanation must reason about identity consistency across images rather
than single-image class logits.

Our attribution-as-measurement framework addresses this gap with
pairwise diagnostics designed for frozen embeddings and calibrated to
biometric operating points.

\subsection{Face Coverage Ratio (FCR) Normalization}
\label{app:fcr}

Faces occupy different fractions of an image; naively ``masking the
face'' in a tight crop removes more pixels (and thus more evidence) than
in a wide shot. We therefore standardize the face budget before any
perturbation by targeting a common face coverage ratio:
\begin{equation}
    \mathrm{FCR}(x)
    = \frac{\mathrm{area}(\text{face mask in } x)}{\mathrm{area}(x)}.
    \label{eq:fcr}
\end{equation}
After resizing/cropping to reach a target FCR, we only compare
perturbations that are equal-area and equal-strength; this lets us
attribute any similarity change to \emph{what} we touched (face vs.\
background), not \emph{how much} we touched. Implementation details
including mask construction and area tolerance are in
Sec.~\ref{app:fca-params} below.

\paragraph{Algorithm.}

\textbf{Input}: Original image, face bounding box $(x_1, y_1, x_2, y_2)$
from RetinaFace detector.

\textbf{Target}: FCR $= 0.33$ (33\% face, 67\% background).

\begin{enumerate}[nosep]
    \item \textbf{Compute face area}:
          $A_{\text{face}} = (x_2 - x_1) \times (y_2 - y_1)$.
    \item \textbf{Determine crop size}:
          $s = \sqrt{A_{\text{face}} / 0.33}$.
    \item \textbf{Center crop}: Face center
          $= ((x_1{+}x_2)/2,\; (y_1{+}y_2)/2)$.
          Crop window:
          $[c_x - s/2,\; c_y - s/2,\; c_x + s/2,\; c_y + s/2]$.
    \item \textbf{Handle boundaries}: If crop extends beyond the image,
          pad with mean color. Track padding ratio:
          $r_{\text{pad}} = \sum_{\text{sides}} \text{pad} / (4s)$.
    \item \textbf{Quality control}: Reject if $r_{\text{pad}} > 0.15$
          (excessive artificial background).
    \item \textbf{Resize}: Crop region to $224 \times 224$
          (or $288 \times 288$ for SSCD).
\end{enumerate}

\textbf{Results}: From 24,000 VGGFace2 images, 23,728 retained (272
rejected for padding). Achieved FCR: mean $= 0.330$, std $= 0.015$.

\paragraph{Face Mask Generation.}
We fit an ellipse to the 5-point landmarks (left eye, right eye, nose,
left mouth, right mouth):
\begin{itemize}[nosep]
    \item \textbf{Center}: Midpoint of (eye center, mouth center).
    \item \textbf{Width axis}: $1.2 \times$ inter-ocular distance.
    \item \textbf{Height axis}: $1.3 \times$ eye-to-mouth distance.
    \item \textbf{Rotation}: Aligned to eye-line angle.
    \item \textbf{Feathering}: 3-pixel Gaussian blur on mask boundary.
\end{itemize}

Binary mask stored as uint8 (255 = face, 0 = background).

\subsection{Face Importance Index (FII)}
\label{app:fii}

If equal-area occlusions to face and background have different effects
on pairwise similarity, the difference is informative about where
identity evidence lives. Let $\mathbf{z}^{\text{bg-occ}}$ and
$\mathbf{z}^{\text{face-occ}}$ denote the $\ell_2$-normalized
embeddings of image $x$ with an equal-area occlusion mask applied to the
background region and face, respectively. For a query--reference pair
$(\mathbf{z}_q, \mathbf{z}_r)$ we measure:
\begin{align}
    \Delta_{\text{face}}
     & = \cos(\mathbf{z}_q, \mathbf{z}_r)
    - \cos\bigl(\mathbf{z}_q^{\text{face-occ}},\;
    \mathbf{z}_r^{\text{face-occ}}\bigr),
    \label{eq:delta-face}                 \\[4pt]
    \Delta_{\text{bg}}
     & = \cos(\mathbf{z}_q, \mathbf{z}_r)
    - \cos\bigl(\mathbf{z}_q^{\text{bg-occ}},\;
    \mathbf{z}_r^{\text{bg-occ}}\bigr),
    \label{eq:delta-bg}
\end{align}
and define $\mathrm{FII} = \Delta_{\text{face}} - \Delta_{\text{bg}}$,
averaged over pairs. Positive FII indicates face-dominant similarity;
negative FII indicates context dominance.

\paragraph{Implementation: equal-area occlusion.}

\textbf{Face occlusion}: Apply Gaussian blur ($\sigma = 8$, kernel
$49 \times 49$) to face region, blended using the face mask with
3-pixel feathering.

\textbf{Background occlusion (equal-area)}: Create an annulus mask
surrounding the face by expanding the face mask outward by $w$ pixels
via morphological dilation, then subtracting the original face mask:
$M_{\text{annulus}} = \mathrm{dilate}(M, w) - M$.
Use binary search to find $w$ such that
$A_{\text{annulus}} = A_{\text{face}} \pm 2\%$.
Apply the same Gaussian blur ($\sigma = 8$) to the annulus region.

Aggregate: $\text{FII}_{\text{mean}} = \text{mean}(\text{FII})$ over
all pairs; 95\% CI via percentile bootstrap (2,000 samples).

\subsection{Context Preference Index (CPI)}
\label{app:cpi}

For each image $i$, we construct two reference images: an
identity-matched image $\mathbf{z}^i_{\text{id}}$ (same person,
different context) and a context-matched image
$\mathbf{z}^i_{\text{ctx}}$ (different person via face-swap, same
context). We apply Gaussian face blur with strength $\sigma$ to all
three images and compute their embeddings. CPI measures how often the
blurred query prefers context over identity:
\begin{equation}
    \mathrm{CPI}(\sigma)
    = \frac{1}{N}\sum_{i=1}^{N}
    \mathbf{1}\!\Bigl[
        \cos\bigl(\mathbf{z}^i_{q,\sigma},\;
        \mathbf{z}^i_{\text{ctx},\sigma}\bigr)
        \;\geq\;
        \cos\bigl(\mathbf{z}^i_{q,\sigma},\;
        \mathbf{z}^i_{\text{id},\sigma}\bigr)
        \Bigr],
    \label{eq:cpi}
\end{equation}
and we report
$\Delta\mathrm{CPI} = \mathrm{CPI}(\sigma_{\max}) -
    \mathrm{CPI}(\sigma_{\min})$ and the crossover $\sigma^*$ when defined.

\paragraph{Implementation: face blur series.}

\textbf{Triplet construction}: For each query image, form:
\begin{itemize}[nosep]
    \item \textbf{SP-DC} (Same Person, Different Context): same identity,
          different scene/clothing.
    \item \textbf{DP-SC} (Different Person, Same Context): face-swapped
          into query's exact background.
\end{itemize}

\textbf{Blur series}: Apply Gaussian blur to face region at
$\sigma \in \{0, 1, 2, 4, 6, 8\}$, blended symmetrically across query,
SP-DC, and DP-SC.

\textbf{Derived metrics}:
\begin{itemize}[nosep]
    \item $\Delta$CPI $= \text{CPI}(\sigma=8) - \text{CPI}(\sigma=0)$:
          Change in context preference.
    \item $\sigma^*$: Crossover blur level where CPI $= 0.5$ (if exists).
\end{itemize}

\subsection{Background Revelation Threshold ($B^*$)}
\label{app:bstar}

As a complementary stress test, we monotonically reveal more background
while keeping the face crop constant and ask when context overtakes
identity. Let $B \in [0,1]$ be the revealed background fraction in a
monotone series that preserves the face crop. For each triplet (query,
identity-matched, context-matched), define:
\begin{equation}
    B^*_{\text{query}}
    = \min\Bigl\{
    B \;:\;
    \cos\bigl(\mathbf{z}_{q,B},\;\mathbf{z}_{\text{ctx}}\bigr)
    \;\geq\;
    \cos\bigl(\mathbf{z}_{q,B},\;\mathbf{z}_{\text{id}}\bigr)
    \Bigr\},
    \label{eq:bstar}
\end{equation}
and report the median across triplets (with censoring at $0$ and $1$).
$B^* \to 0$ indicates context dominance (minimal background suffices to
overtake identity); $B^* \to 1$ indicates face dominance (identity
resists background revelation). Implementation parameters are in
Sec.~\ref{app:fca-params} below.

\subsection{Implementation Parameters and Quality Controls}
\label{app:fca-params}

\paragraph{Blur parameters.}
All Gaussian blurs use kernel size $= 2 \times \lceil 3\sigma \rceil + 1$.
For $\sigma = 8$ (heavy occlusion), kernel $= 49 \times 49$.

\paragraph{Face-swap quality.}
DP-SC pairs generated via INSwapper (InsightFace):
\begin{itemize}[nosep]
    \item Success rate: 98.0\% (23,254 / 23,728 attempted).
    \item Poisson blending with 3-pixel feather.
    \item Histogram matching for color consistency.
\end{itemize}
Visual inspection: $< 2\%$ show minor artifacts (acceptable for
stress-test purposes).

\paragraph{FCR matching.}
For SP-DC pairs, FCR matched within $\pm 2\%$ tolerance.
Actual FCR difference: mean $= 0.70\%$, max $= 2.0\%$.

\paragraph{Equal-area tolerance.}
FII background annulus matched to face area within $\pm 2\%$.
Actual area difference: mean $= 0.8\%$, max $= 2.0\%$.

\paragraph{Monotonicity checks.}
For $B^*$, we verify:
\begin{itemize}[nosep]
    \item $\text{Sim}_{\text{DPSC}}(B)$ increases with $B$
          (94\% of triplets).
    \item $\text{Sim}_{\text{SPDC}}(B)$ approximately flat
          (98\% of triplets).
\end{itemize}
Non-monotonic cases ($< 6\%$) attributed to noise in the 5-point
$B$-level grid.

\clearpage

\section{Non-Linear Robustness: MLP Verifier Results}
\label{app:mlp}

To test whether identity information persists for moderate non-linear
attackers after ISP, we train a projection-only MLP verifier under the
same open-set, identity-disjoint protocol as the Ridge probe.
The MLP (two hidden layers, ReLU) operates on cosine similarities
between query and support embeddings; all hyperparameters and thresholds
are frozen on $A_{\text{val}}$ before evaluation on $A_{\text{test}}$.
FR models (ArcFace, AdaFace) are included as pre-ISP baselines only;
ISP is not applied to FR encoders (marked ``---'').

Table~\ref{tab:mlp_verifier_full} reports TAR@FAR$=10^{-4}$ (\%) for
$k \in \{1, 4, 16\}$ shots on both CelebA-20 and VGGFace2-20.
Non-FR encoders already exhibit low non-linear leakage in raw
embeddings. After ISP, TAR drops further, with DINOv2/v3/SSCD reaching
near zero across all $k$ on both datasets. CLIP, with its larger embedding
dimension and broader identity subspace, retains modest residual leakage
after ISP, consistent with the rank-sweep results in
Table~\ref{tab:rank_sweep}.
FR models achieve high pre-ISP TAR on both datasets, confirming
the MLP baseline is well-calibrated; ISP is not evaluated on FR encoders.

\begin{table}[h]
    \centering
    \small
    \setlength{\tabcolsep}{3.5pt}
    \caption{%
    \textbf{Projection-only MLP verifier (CelebA-20).}
    TAR@FAR$=10^{-4}$ (\%).
    Pre-ISP uses raw embeddings; Post-ISP applies within-dataset ISP-W.
    FR models are shown pre-ISP only (ISP not evaluated on FR encoders).
    }
    \label{tab:mlp_verifier_full}
    \begin{tabular}{@{}ll ccc ccc@{}}
        \toprule
        \textbf{Type} & \textbf{Model}
                      & \multicolumn{3}{c}{Pre-ISP} & \multicolumn{3}{c}{Post-ISP} \\
        \cmidrule(lr){3-5}\cmidrule(lr){6-8}
                      &         & $k{=}1$ & $k{=}4$ & $k{=}16$
                               & $k{=}1$ & $k{=}4$ & $k{=}16$ \\
        \midrule
        \multirow{4}{*}{\rotatebox{90}{\small Non-FR}}
                      & DINOv2  & 1.4  & 2.2  & 3.8  & 0.0 & 0.8 & 0.4 \\
                      & DINOv3  & 2.2  & 3.4  & 5.9  & 0.5 & 0.3 & 0.9 \\
                      & SSCD    & 2.4  & 3.9  & 5.8  & 0.8 & 1.1 & 0.7 \\
                      & CLIP    & 18.3 & 22.9 & 26.9 & 1.6 & 1.8 & 0.5 \\
        \midrule
        \multirow{2}{*}{\rotatebox{90}{\small FR}}
                      & ArcFace & 91.9 & 90.9 & 92.8 & \multicolumn{3}{c}{---} \\
                      & AdaFace & 92.2 & 91.2 & 93.0 & \multicolumn{3}{c}{---} \\
        \bottomrule
    \end{tabular}
\end{table}

\begin{table}[h]
    \centering
    \small
    \setlength{\tabcolsep}{3.5pt}
    \caption{%
    \textbf{Projection-only MLP verifier (VGGFace2-20).}
    TAR@FAR$=10^{-4}$ (\%).
    Pre-ISP uses raw embeddings; Post-ISP applies within-dataset ISP-W.
    FR models are shown pre-ISP only (ISP not evaluated on FR encoders).
    }
    \label{tab:mlp_verifier_vgg}
    \begin{tabular}{@{}ll ccc ccc@{}}
        \toprule
        \textbf{Type} & \textbf{Model}
                      & \multicolumn{3}{c}{Pre-ISP} & \multicolumn{3}{c}{Post-ISP} \\
        \cmidrule(lr){3-5}\cmidrule(lr){6-8}
                      &         & $k{=}1$ & $k{=}4$ & $k{=}16$
                               & $k{=}1$ & $k{=}4$ & $k{=}16$ \\
        \midrule
        \multirow{4}{*}{\rotatebox{90}{\small Non-FR}}
                      & DINOv2  & 0.2 & 0.4 & 0.9 & 0.0 & 0.0 & 0.0 \\
                      & DINOv3  & 0.2 & 0.7 & 0.9 & 0.0 & 0.0 & 0.0 \\
                      & SSCD    & 0.0 & 0.0 & 0.0 & 0.0 & 0.0 & 0.0 \\
                      & CLIP    & 1.3 & 2.4 & 2.3 & 0.0 & 0.0 & 0.0 \\
        \midrule
        \multirow{2}{*}{\rotatebox{90}{\small FR}}
                      & ArcFace & 89.3 & 89.3 & 89.3 & \multicolumn{3}{c}{---} \\
                      & AdaFace & 88.8 & 88.8 & 88.8 & \multicolumn{3}{c}{---} \\
        \bottomrule
    \end{tabular}
\end{table}

\newpage
\section{DISC2021 Copy-Detection Utility (SSCD)}
\label{app:disc}

Table~\ref{tab:utility_disc_raw} reports Recall@$k$ retained after
projection on the DISC2021 copy-detection benchmark for SSCD.
Both ISP and LEACE preserve over 95\% of baseline recall across all
$k$, confirming that task-specific retrieval utility is largely
unaffected.

\begin{table}[t]
    \centering
    \caption{\textbf{DISC2021 copy-detection (SSCD).}
        Percentage of Recall@$k$ retained after projection,
        relative to unprojected embeddings.}
    \label{tab:utility_disc_raw}
    \small
    \begin{tabular}{lccc}
        \toprule
        \textbf{Metric} & \textbf{ISP} & \textbf{LEACE} \\
        \midrule
        Recall@1        & 95.0\%       & 95.1\%         \\
        Recall@5        & 96.5\%       & 96.4\%         \\
        Recall@10       & 96.0\%       & 96.1\%         \\
        \bottomrule
    \end{tabular}
\end{table}

\clearpage

\section{Subspace Removal: Extended Background}
\label{app:subspace-removal}

There is a mature literature on removing sensitive information from fixed
representations via linear subspace editing. Two broad families dominate.

\paragraph{Classifier-driven nulling.}
Iterative or minimax procedures train a linear adversary for a protected
attribute and project onto (the intersection of) its nullspaces.
INLP~\cite{Ravfogel2020} repeatedly fits a linear classifier, projects
the data onto its nullspace, and iterates until the attribute is no
longer linearly predictable. RLACE~\cite{Ravfogel2022RLACE} casts the
same goal as a minimax program and solves it in closed form, yielding a
rank-constrained projection that minimizes worst-case linear
leakage. Both families aim for strong linear unpredictability guarantees
but require repeated classifier fitting (or solving a minimax program)
and can rotate or compress the feature geometry in ways that are
difficult to audit post hoc.

\paragraph{Moment- and covariance-driven nulling.}
One-shot transforms estimated from class means and scatter matrices
offer a complementary approach. SAL~\cite{Shao2023SAL} removes guarded
attribute information via spectral decomposition of class-conditioned
representations. LEACE~\cite{Belrose2023LEACE} derives a closed-form
least-squares erasure that removes all linearly accessible information
about a concept while minimizing distortion to the embedding, and comes
with a formal optimality certificate under squared loss. Related
fair-PCA methods~\cite{Kamani2020FairPCA} optimize Pareto trade-offs
between utility and fairness without enforcing zero leakage, offering a
softer alternative when complete erasure is unnecessary.

These moment-based methods explicitly target between-class mean
structure (often in a whitened Fisher geometry) and are attractive for
deployment because they require no adversary training and produce a
fixed, exportable projection matrix.

\paragraph{Positioning of ISP.}
Prior concept-erasure work largely targets binary or low-cardinality
attributes (e.g., gender, sentiment) and reports accuracy drops on
supervised classification tasks. In contrast, facial identity in our
setting is high-cardinality and open-set, and risk must be quantified at
low FAR under disjoint identities. It is therefore unclear a priori
whether the identity signal in non-FR embeddings: (a)~concentrates in a
compact, transferable subspace, (b)~can be linearly ``certified away''
without harming non-biometric utility, and (c)~remains suppressed under
stronger, projection-only non-linear probes. Our study directly answers
these questions with attacker-aware metrics (open-set TAR@low-FAR,
projection-only MLP, template inversion) and with cross-dataset transfer
and robustness checks, which are typically absent from concept-erasure
evaluations.

ISP adopts a moment-based, one-shot design: compute per-identity mean
differences, take the SVD, and project onto the orthogonal complement of
the top-$r$ ``identity'' directions; optionally whiten to obtain a
Fisher-space certificate~\cite{Belrose2023LEACE}. Under
homoscedasticity, this removes between-class mean structure and yields a
clear linear leakage certificate while preserving the complementary
subspace. Compared to iterative/minimax methods (INLP/RLACE), ISP is:
\begin{enumerate}[nosep]
    \item \emph{Auditable}: rank $r$ directly controls the
          privacy-utility trade-off and enables energy diagnostics.
    \item \emph{Lightweight}: a single SVD with no adversary training or
          hyperparameter grids.
    \item \emph{Model-agnostic and exportable}: a fixed $P$ that plugs
          into any retrieval pipeline.
    \item \emph{Deployment-friendly}: stable, deterministic, no
          retraining, sub-millisecond latency.
\end{enumerate}
In our evaluations, this simple construction is sufficient to drive
linear identity accessibility near chance while retaining most utility,
and its fixed projector generalizes across datasets - properties that
alternatives must match to be practical at scale.

We emphasize that linear subspace removal certifies against linear
attackers; non-linear leakage may persist. In exchange, linear methods
typically preserve utility far better than end-to-end adversarial
training and serve as auditable, deployable baselines that stronger
mitigations must beat. We provide empirical evidence on non-linear
robustness via a projection-only MLP verifier in
Appendix~\ref{app:mlp} and Sec.~4.2.

\clearpage
\section{Attribute-Swap Identity Inference}
\label{app:attrswap}

Prior work has shown that identity can remain stable under semantic
attribute changes (e.g., gender) in FR embeddings. A natural question
is whether this holds for non-FR encoders after ISP.
We evaluate PCA-based attribute-swap attacks that attempt to modify
the gender attribute while preserving identity post-ISP, following the
manipulation strategy of Kim et al.\ (NeurIPS '25).

\subsection{Protocol}
\label{app:attrswap-protocol}

\paragraph{Setup.}
We work with CelebA-20 embeddings ($N = 500$ victims per direction).
A linear gender classifier is trained on training-split embeddings;
its decision boundary normal defines the gender direction
$\mathbf{v}_{\text{gender}}$.
PCA is used to find the top-$k$ directions in the subspace orthogonal
to within-identity variation that maximally covary with the gender label.
A victim embedding $z$ is projected along these $k$ directions to produce
a manipulated embedding $\tilde{z}$.

\paragraph{Metrics.}
We report three quantities at $k = 128$ components:
\begin{itemize}[nosep]
    \item \textbf{Accept}: fraction of victims where
          $\cos(z, \tilde{z}) \geq \tau$ (identity similarity threshold
          at FAR $= 10^{-3}$), i.e.\ the manipulated embedding is still
          identity-proximate.
    \item \textbf{Flip}: fraction of victims where the gender classifier
          predicts the \emph{target} gender for $\tilde{z}$.
    \item \textbf{Joint}: fraction of victims satisfying \emph{both}
          Accept and Flip simultaneously---the attacker's true success rate.
\end{itemize}

\newpage
\subsection{Results}
\label{app:attrswap-results}

\begin{table}[h]
    \centering
    \caption{Attribute-swap identity inference at $k=128$.
        \textbf{Joint} = fraction of victims where identity is preserved
        (sim $\geq \tau_{\mathrm{FAR}=10^{-3}}$) \emph{and} gender is
        flipped. FR models allow highly reliable joint manipulation;
        non-FR encoders do not. 500 victims per direction on CelebA-20.}
    \label{tab:suppl_attrswap}
    \small
    \begin{tabular}{l l c c c}
        \toprule
        \textbf{System}       & \textbf{Direction}
                              & \textbf{Accept}         & \textbf{Flip} & \textbf{Joint}                   \\
        \midrule
        DINOv2                & $\text{F}{\to}\text{M}$ & 96.8\%        & 14.6\%         & 14.0\%          \\
        \quad + ISP ($r=192$) & $\text{F}{\to}\text{M}$ & 96.2\%        & 30.2\%         & 28.8\%          \\
        DINOv2                & $\text{M}{\to}\text{F}$ & 98.4\%        & 14.6\%         & 14.4\%          \\
        \quad + ISP ($r=192$) & $\text{M}{\to}\text{F}$ & 98.2\%        & 26.6\%         & 26.2\%          \\
        \midrule
        DINOv3                & $\text{F}{\to}\text{M}$ & 99.6\%        & 13.0\%         & 12.8\%          \\
        \quad + ISP ($r=192$) & $\text{F}{\to}\text{M}$ & 100.0\%       & 17.8\%         & 17.8\%          \\
        DINOv3                & $\text{M}{\to}\text{F}$ & 100.0\%       & 10.2\%         & 10.2\%          \\
        \quad + ISP ($r=192$) & $\text{M}{\to}\text{F}$ & 100.0\%       & 17.4\%         & 17.4\%          \\
        \midrule
        CLIP                  & $\text{F}{\to}\text{M}$ & 100.0\%       & 9.2\%          & 9.2\%           \\
        \quad + ISP ($r=192$) & $\text{F}{\to}\text{M}$ & 98.0\%        & 13.8\%         & 13.4\%          \\
        CLIP                  & $\text{M}{\to}\text{F}$ & 100.0\%       & 6.6\%          & 6.6\%           \\
        \quad + ISP ($r=192$) & $\text{M}{\to}\text{F}$ & 98.6\%        & 12.0\%         & 11.8\%          \\
        \midrule
        SSCD                  & $\text{F}{\to}\text{M}$ & 100.0\%       & 37.0\%         & 37.0\%          \\
        \quad + ISP ($r=192$) & $\text{F}{\to}\text{M}$ & 100.0\%       & 35.0\%         & 35.0\%          \\
        SSCD                  & $\text{M}{\to}\text{F}$ & 100.0\%       & 23.6\%         & 23.6\%          \\
        \quad + ISP ($r=192$) & $\text{M}{\to}\text{F}$ & 100.0\%       & 29.0\%         & 29.0\%          \\
        \midrule
        \multicolumn{5}{l}{\textit{Face Recognition Models (no ISP variant)}}                              \\
        \midrule
        ArcFace               & $\text{F}{\to}\text{M}$ & 99.8\%        & 81.2\%         & \textbf{81.2\%} \\
        ArcFace               & $\text{M}{\to}\text{F}$ & 99.8\%        & 73.6\%         & \textbf{73.6\%} \\
        AdaFace               & $\text{F}{\to}\text{M}$ & 99.8\%        & 70.4\%         & \textbf{70.4\%} \\
        AdaFace               & $\text{M}{\to}\text{F}$ & 99.6\%        & 52.2\%         & \textbf{52.2\%} \\
        \bottomrule
    \end{tabular}
\end{table}

\end{document}